\documentclass[a4paper,twocolumn,11pt,unpublished]{quantumarticle}
\pdfoutput=1
\usepackage[utf8]{inputenc}
\usepackage[english]{babel}
\usepackage[T1]{fontenc}
\usepackage{amsmath}

\usepackage{graphicx} 
\usepackage{subcaption} 

\usepackage{booktabs}

\usepackage[numbers,sort&compress]{natbib}
\usepackage{hyperref}

\begin{document}

\title{Learning to Rank Tensor Network Contraction Plans for GPU-Accelerated Quantum Circuit Simulation}

\author{Alfred M. Pastor}
\email{alfred.pastor@uv.es}
\affiliation{Department of Computer Science, Universitat de València, 46100 Burjassot, Spain}
\orcid{0000-0002-7740-6354}
\author{Maribel Castillo}
\email{castillo@uji.es}
\affiliation{Department of Computer Science and Engineering, Universitat Jaume I, 12071 Castelló de la Plana, Spain}
\orcid{0000-0002-2826-3086}
\author{Jose M. Badia}
\email{badia@uji.es}
\affiliation{Department of Computer Science and Engineering, Universitat Jaume I, 12071 Castelló de la Plana, Spain}
\orcid{0000-0002-5927-0449}
\maketitle


\begin{abstract}
Classical simulation remains essential for developing and validating quantum algorithms, but its cost grows rapidly with circuit size. Tensor-network contraction can reduce this cost by exploiting circuit structure, although its efficiency depends strongly on the chosen contraction plan. On GPUs, plans with similar theoretical complexity may perform very differently because execution also depends on parallelism, reduction structure, memory traffic, and contraction geometry.

We present a learning-to-rank framework for selecting efficient contraction plans before executing them. Each plan is represented by structural features derived directly from its sequence of pairwise contractions, and gradient-boosted rankers are trained from GPU measurements using listwise and pairwise objectives. We evaluate the resulting models on diverse circuit families, using separate in-distribution and circuit-family-shift test sets, and compare them with random and MinFill-based baselines.

The learned rankers generally identify better plans, with the listwise model providing the strongest overall decision quality. We also study backend shift by comparing empirical plan orderings on two GPU architectures and evaluating the source-trained models on the second device without retraining. The rankings remain substantially, though not perfectly, stable across GPUs, and the models retain useful decision quality. These results support Learning to Rank as a practical way to reduce contraction-plan search, while showing that performance remains partly backend dependent.
\end{abstract}


\section{Introduction}
\label{sec:intro}

Classical simulation remains an important part of quantum-computing research. It supports the design and testing of algorithms, the validation of experiments, and the benchmarking of quantum hardware. State-vector simulation is general and conceptually simple, but its memory requirements grow exponentially with the number of qubits. Tensor-network methods provide an alternative when the circuit has exploitable structure, allowing some simulations to go beyond the sizes that can be handled by storing the complete quantum state~\cite{MaS08, Oru14, GrK21}.

The cost of a tensor-network simulation depends strongly on the order in which its tensors are contracted. A contraction plan expresses that order as a sequence of pairwise contractions, and different plans for the same network can have very different computational and memory requirements~\cite{PHV14}. Finding an optimal order is NP-hard, so practical tools rely on heuristics such as MinFill, graph partitioning, community detection, and treewidth-oriented search~\cite{BDK00, GiN02, HaS18, GrK21}. These methods produce useful candidates, but a plan that appears favorable according to a theoretical cost estimate is not necessarily the one that runs fastest on the target hardware.

This distinction is particularly relevant on GPUs. Pairwise tensor contractions can often be reduced to tensor permutations and matrix multiplications, operations that make effective use of GPU parallelism. Their execution time, however, is not determined by floating-point count alone. Output dimensions, reduction size, contraction geometry, memory movement, kernel granularity, and the balance between many small operations and a few dominant contractions can all affect performance~\cite{Cor24, Cor25}. Plans with similar theoretical complexity may therefore behave differently when executed through a particular GPU and tensor-library stack.

The practical problem considered here is simple to state: given a quantum circuit and several candidate contraction plans, select the plan expected to minimize execution time, or a short list containing a near-optimal candidate on the GPU. Measuring all of them would reveal the fastest one, but it could require a substantial fraction of the computation that the selection stage is intended to save. In many settings, an exact runtime estimate is also unnecessary. The relevant decision is which plan should be executed, or which small set of plans is worth testing.

We formulate this task as a grouped Learning to Rank (LTR) problem. The candidates associated with one circuit form a ranking group, and the model assigns each plan a score used only to order the plans within that group. The intended outputs are either a direct Top-1 choice or a short Top-\(k\) list from which the best plan can be selected after a limited number of executions. LTR is designed for this type of ordinal decision and has been used in quantum computing for related selection problems, including the ranking of logically equivalent circuit layouts for execution on quantum hardware~\cite{Liu09, HMH24}. Here, the ranked objects are tensor-network contraction plans and the target ordering is determined by their measured GPU performance.

Each plan is represented by structural descriptors computed from its sequence of pairwise contractions. They summarize properties related to computational load, output-side parallelism, reduction structure, contraction geometry, and plan granularity. The representation requires neither preliminary execution nor profiling of the candidate plans. It can therefore be evaluated before the expensive tensor contractions are carried out.

The labels used for training are obtained from measured execution times. For each circuit, the candidates are evaluated through their speedup relative to the corresponding MinFill plan, and this value is transformed into a graded relevance label. The inputs are derived only from the contraction structure, but the labels reflect a particular GPU and software stack. The resulting model is therefore not assumed to be hardware independent: backend dependence can enter through the supervision signal even when the backend is not explicitly encoded among the features.

We train gradient-boosted rankers with XGBoost~\cite{ChG16}, using its Normalized Discounted Cumulative Gain (NDCG) and pairwise ranking objectives. The dataset contains 225 circuit groups and seven candidate plans per circuit, generated with several contraction-order heuristics and measured on an NVIDIA RTX A6000. All plans belonging to the same circuit are kept together during partitioning and validation. Model development is followed by evaluation on a locked in-distribution (ID) test set and on an out-of-distribution (OOD) set containing the complete quantum Fourier transform (QFT)-based circuit family, comprising QFT circuits and QFT-derived benchmark variants. This family is entirely excluded from training and validation. The models are compared with random and MinFill-based ranking strategies using Top-1, Top-3, regret, and win rate.

The NDCG-oriented model includes the fastest plan among its first three recommendations for \(96\%\) of the circuits in the locked in-distribution test, with low decision regret. When the complete QFT-based family is withheld from development, its Top-3 rate falls to \(62.9\%\). The model therefore retains useful information for a circuit family it has not seen, but the loss under this form of domain shift is clear.

We also study dependence on the execution backend. The same candidate plans are measured on an NVIDIA Tesla V100, and the analysis separates two questions. First, we compare the two empirical ground-truth rankings without involving a learned model; the GPUs select the same fastest plan in \(84\%\) of the 225 circuit groups. Second, we keep the Ampere-trained models and their predicted rankings fixed, but evaluate them on the locked test set using labels recomputed from the Volta measurements. Under this zero-shot backend shift, the NDCG-oriented model retains a Top-3 rate of \(92\%\). These results show useful transfer for the two NVIDIA architectures considered, but they do not establish general hardware portability.

The current evidence is limited to seven candidates per circuit, one held-out circuit family, and two GPUs supported by closely related software stacks. Within that scope, the experiments indicate that structural information from a contraction plan is sufficient to support useful ranking decisions, that an NDCG-oriented objective gives the best balance of Top-\(k\) accuracy and decision regret in the present dataset, and that part of the learned ordering survives both circuit-family and backend shift.

The main contributions of this work are:

\begin{itemize}
\item We formulate GPU contraction-plan selection as a grouped Learning to Rank problem, aimed at choosing either one candidate or a short list of candidates before execution.

\item We define a fixed-size structural representation of contraction plans that captures GPU-relevant aspects of work, parallelism, reduction, geometry, and granularity without requiring timing measurements at inference time.

\item We develop a group-aware training and evaluation protocol for NDCG-oriented and pairwise XGBoost rankers, using both ranking accuracy and decision-oriented regret on locked in-distribution and circuit-family-shift test sets.

\item We examine backend dependence through two separate analyses: the stability of the empirical plan rankings across Ampere and Volta, and the zero-shot evaluation of fixed Ampere-trained models using labels measured on Volta.
\end{itemize}

The remainder of the paper is organized as follows. Section~\ref{sec:rela} reviews related uses of machine learning and LTR in quantum computing. Section~\ref{sec:back} introduces tensor-network simulation, GPU execution of contraction plans, candidate-generation heuristics, and gradient-boosted ranking. Section~\ref{sec:feat_eng} presents the structural feature representation, with further details in Appendix~\ref{app:feature_details}. Section~\ref{sec:meth} describes data collection, grouped validation, feature selection, model optimization, and the experimental platform. Section~\ref{sec:exper} reports the learning-curve, in-distribution, and circuit-family-shift results. Section~\ref{sec:cross} studies ground-truth ranking stability and zero-shot model transfer across GPUs. Section~\ref{sec:conc} presents the conclusions and planned extensions.

\section{Related Work}
\label{sec:rela}

The intersection of artificial intelligence (AI) and quantum computing is a bidirectional research frontier. On one hand, quantum computing offers the potential to accelerate machine learning tasks, a field often referred to as Quantum Machine Learning (QML). Recent surveys have extensively covered the landscape of QML, highlighting algorithms such as quantum support vector machines, quantum neural networks, and variational quantum classifiers, along with their potential advantages and current limitations on near-term quantum hardware~\cite{ScP21, LKP23}. On the other hand, classical machine learning is increasingly being employed to tackle core challenges in quantum computing. This includes using neural networks for quantum error correction and decoding, as exemplified by the AlphaQubit recurrent-transformer architecture, which learns to predict and correct errors in quantum processors~\cite{CNA24}. Such work underscores the growing synergy between the two fields, where classical learning algorithms help advance quantum hardware and simulation capabilities.

Beyond quantum error correction, various machine learning techniques have been applied to quantum computing challenges. For instance, support vector machines (SVMs) have been used for quantum state tomography~\cite{KBR20} and to classify quantum phases of matter~\cite{WZW19}. Supervised learning has also been used to predict suitable device, compiler, and configuration choices for quantum-circuit compilation, reducing the need for exhaustive exploration~\cite{CLW21}. Neural networks have been trained to learn the structure of quantum states~\cite{TMC18}, and to guide variational quantum eigensolvers~\cite{VBB19}. 

One particularly suitable paradigm for candidate selection is LTR, a family of supervised learning techniques designed to optimize the ordering of a list of items rather than predicting their absolute values. LTR methods are widely used in information retrieval, web search, recommendation systems, and natural language processing~\cite{Liu09, Mah23}. In recent years, LTR has also been successfully applied to a diverse range of problems, ranging from e-commerce product search~\cite{KHM24, GXK26} to search engine optimization and document retrieval~\cite{BLN23, GJH23, Wan22}.

Only a few studies have applied LTR techniques to quantum computing problems. The most relevant to our research is the work by Li et al.~\cite{LZY25}, who proposed QCDeploy, an LTR-based method to rank different deployment strategies for quantum circuits on serverless platforms. Another related study by Hartnett et al.~\cite{HMH24} introduced an LTR-based approach to rank logically equivalent quantum circuits for hardware-aware layout selection on IBM processors. These studies demonstrate the potential of LTR for automating optimization decisions in quantum computing. 

To the best of our knowledge, prior work has not applied LTR to
tensor-network contraction or to the prediction of contraction-plan
efficiency on GPUs. Motivated by this gap, the next section presents an LTR-based framework that ranks candidate contraction plans and identifies the one expected to achieve the fastest execution on modern GPUs, with the aim of reducing the cost of contraction-plan selection in large-scale quantum circuit simulation.

\section{Background}
\label{sec:back}

\subsection{Tensor Networks for Quantum Circuit Simulation}

Tensor networks provide a powerful framework for representing quantum many-body states through networks of interconnected tensors~\cite{Oru14}. Quantum circuit simulation can be formulated as the contraction of a tensor network, where each quantum gate is represented by a tensor whose indices correspond to the input and output qubits on which it acts~\cite{Oru14, LBC25}. Computing probability amplitudes or observables amounts to contracting the internal indices of the resulting network.

Compared with full state-vector simulation, tensor networks exploit local entanglement to alleviate the exponential growth of memory requirements~\cite{MaS08}. Their efficiency, however, depends strongly on the contraction plan used to evaluate the network. A contraction plan defines the sequence in which tensors are contracted and therefore determines both the computational cost and the memory footprint of the simulation~\cite{MaS08}. Since finding an optimal contraction order is NP-hard, practical implementations rely on heuristic methods and must also contend with overheads associated with tensor transpositions and index reordering~\cite{Ogo19}.

\subsection{GPU Execution of Contraction Plans}

In practice, tensor contractions can be decomposed into tensor permutations, reshaping operations, and matrix multiplications~\cite{SpB18, Mat18}. Consequently, matrix products dominate the arithmetic cost, whereas tensor permutations and memory movement constitute the main sources of overhead.

These operations map naturally onto GPU architectures. State-of-the-art simulators such as cuQuantum~\cite{Cor24} and qFlex~\cite{VBN19} exploit optimized libraries, including cuBLAS and cuTENSOR, to accelerate tensor network simulation. However, execution time depends not only on the number of floating-point operations (FLOPs), but also on memory traffic, contraction geometry, intermediate tensor sizes, and the amount of exposed parallelism. Consequently, contraction plans with similar asymptotic costs may exhibit substantially different runtimes on the same hardware.

\subsection{Candidate Contraction Plans}

A wide variety of heuristics have been proposed to address the contraction-order optimization problem. Although they all pursue the same objective, reducing computational cost and memory consumption, they differ in the criteria used to guide the search and the graph structures they exploit. Consequently, these methods can be broadly grouped into several families according to their underlying principles~\cite{GrK21}.
Each family emphasizes different trade-offs between computational cost, memory requirements, and graph structure, often producing contraction plans with similar asymptotic costs but different execution characteristics.

A first family consists of \textbf{greedy methods}, which iteratively select contractions according to local criteria. Depending on the objective, these heuristics may prioritize minimizing FLOPs~\cite{GrK21, Ogo19}, intermediate tensor sizes~\cite{DFG18}, or balancing both factors. Although computationally inexpensive, their local nature may prevent them from finding globally optimal orders.

A second family is based on \textbf{graph elimination and treewidth heuristics}. Methods such as MinFill and Min-Degree aim to reduce the effective treewidth of the network, which is closely related to contraction complexity~\cite{BDK00, DFG18}.

Another class relies on \textbf{graph partitioning and separator techniques}. Algorithms such as FlowCutter and community-detection approaches exploit the graph structure to construct efficient contraction orders with favorable memory requirements and lower effective widths~\cite{HaS18, Str17, GiN02, NeG04}.

Finally, \textbf{hypergraph partitioning techniques}, as employed by \textit{cotengra}~\cite{GrK21}, recursively partition the network and can be combined with slicing strategies to further reduce memory consumption. These approaches are among the most effective for large and complex tensor networks.

Although these families often generate contraction plans with similar asymptotic costs, their execution times may differ substantially on GPU architectures. Consequently, selecting the most efficient plan among a set of candidates becomes a decision problem. Rather than generating new contraction orders, the objective of this work is to rank a pool of candidate plans and identify those most likely to achieve the best execution performance on the target GPU.

\subsection{Learning to Rank with Gradient Boosted Trees}

\subsubsection{Learning to Rank}

LTR refers to a family of supervised learning techniques designed to optimize the ordering of items rather than their absolute values~\cite{Liu09}. LTR methods are commonly categorized into three paradigms: \textit{pointwise}, \textit{pairwise}, and \textit{listwise}. Pointwise methods treat ranking as a regression or classification problem on individual items, pairwise methods learn relative preferences between pairs of items, and listwise methods directly optimize ranking-quality metrics over the entire list.

Since the objective of this work is to identify the best contraction plans among a set of candidates rather than predict their exact execution times, LTR provides a natural framework for contraction-plan selection. In particular, practical usage emphasizes ranking quality at the top positions, where selecting the best or near-best plans is more important than accurately modeling the entire ordering.

\subsubsection{Gradient-Boosted Trees for Ranking}

Gradient-boosted trees constitute one of the most effective approaches for LTR tasks. In this work, we employ XGBoost~\cite{ChG16}, which combines multiple decision trees in an additive model and provides efficient implementations of ranking objectives with built-in regularization. XGBoost is particularly well suited for structured tabular data and can capture complex non-linear relationships while maintaining computational efficiency.

XGBoost supports both pairwise and listwise ranking objectives. The \texttt{rank:pairwise} objective, rooted in RankNet~\cite{BSR05}, learns relative preferences between pairs of items and seeks to minimize ranking inversions. In contrast, the \texttt{rank:ndcg} objective implements LambdaMART~\cite{Bur10}, directly optimizing ranking quality through NDCG. Because NDCG emphasizes the top positions of the ranking, it naturally aligns with our goal of selecting reliable Top-1 and Top-3 contraction plans.

Like other machine learning models, XGBoost's performance depends on a set of hyperparameters controlling tree complexity, regularization, and learning behavior. Table~\ref{tab:xgboost_hyperparams} summarizes the most important hyperparameters for ranking tasks, while the hyperparameter optimization procedure and the selected configurations are described in Section~\ref{sec:meth}.

\begin{table*}[t]
\centering
\small
\caption{Main hyperparameters of XGBoost for learning-to-rank.}
\label{tab:xgboost_hyperparams}
\begin{tabular}{llp{6cm}}
\toprule
\textbf{Parameter} & \textbf{Typical Value} & \textbf{Description} \\
\midrule
\texttt{objective} & \texttt{rank:ndcg} & Loss function; \texttt{rank:ndcg} optimizes NDCG directly, while \texttt{rank:pairwise} minimizes the number of misordered pairs~\cite{BSR05, Bur10}. \\
\texttt{eval\_metric} & \texttt{ndcg@k} & Evaluation metric; NDCG truncated at position $k$. \\
\texttt{eta} & 0.01--0.1 & Learning rate (shrinkage); smaller values require more trees but improve generalization. \\
\texttt{max\_depth} & 3--10 & Maximum depth of each tree; deeper trees capture more complex interactions but increase overfitting risk. \\
\texttt{subsample} & 0.5--1.0 & Fraction of training instances sampled per tree; adds stochasticity and reduces overfitting. \\
\texttt{colsample\_bytree} & 0.5--1.0 & Fraction of features sampled per tree; similar to \texttt{subsample} but for columns. \\
\texttt{lambda} & 0--10 & L2 regularization term on leaf weights; larger values penalize large weights. \\
\texttt{gamma} & 0--5 & Minimum loss reduction required to split a node; larger values lead to more conservative trees. \\
\texttt{min\_child\_weight} & 1--10 & Minimum sum of instance weights in a child node; prevents overfitting on rare patterns. \\
\bottomrule
\end{tabular}
\end{table*}

The choice of these parameters strongly influences the trade-off between bias and variance. For ranking tasks, a moderate learning rate (e.g., $\eta=0.01$), shallow to medium trees (e.g., $\texttt{max\_depth}=6$), and subsampling (e.g., $\texttt{subsample}=0.7$) are commonly employed to avoid overfitting while maintaining high ranking accuracy.

\section{Feature Engineering}
\label{sec:feat_eng}

The contraction path is the dominant algorithmic degree of freedom in exact tensor-network simulation: changing only the order of pairwise contractions can alter both runtime and memory footprint by several orders of magnitude. Classical width-based reasoning explains part of this variability, but it is not sufficient for our setting. In GPU-accelerated contraction, runtime depends not only on the amount of work, but also on output size, reduction structure, contraction geometry, and the amount of exposed parallelism. Our goal is therefore to construct a feature space that is computable directly from a contraction plan before execution, yet expressive enough to capture the structural factors that most influence measured runtime.

\subsection{Cost-model overview}

Consider the \(i\)-th pairwise contraction step of a plan,
\begin{equation}
O_i = \mathrm{contract}_{S_i}(A_i,B_i),
\end{equation}
where \(S_i\) is the set of shared indices contracted between tensors \(A_i\) and \(B_i\). For binary tensor networks, it is natural to work in base-2 logarithmic coordinates. Let
\begin{align}
k_i &= |S_i|,\\
m_i &= \mathrm{rank}(A_i)-k_i,\\
n_i &= \mathrm{rank}(B_i)-k_i.
\end{align}
After a suitable permutation of modes, every pairwise contraction can be written in general matrix multiplication (GEMM)-like form,
\begin{equation}
\begin{split}
(M \times K) \cdot (K \times N) \rightarrow (M \times N),\\
\qquad
M=2^{m_i},\; K=2^{k_i},\; N=2^{n_i},
\end{split}
\label{eq:gemm_view}
\end{equation}
This representation provides a compact abstraction of the local contraction step, since it makes explicit the three structural ingredients that most strongly influence execution: the scale of the local work, the output size, and the reduction dimension.

\subsection{Step-level primitives}

From this local model, we define four step-level primitives:
\begin{align}
c_i &= m_i + n_i + k_i,\\
p_i &= m_i + n_i,\\
k_i &= |S_i|,\\
d_i &= \frac{|m_i-n_i|}{m_i+n_i+\varepsilon},
\qquad \varepsilon \ll 1.
\end{align}

These four quantities form the irreducible basis of our feature space. The exponent \(c_i\) is a proxy for the local computational burden and scales with the equivalent FLOP count. The exponent \(p_i\) captures the output size of the step and therefore acts both as a proxy for local memory pressure and for output-side data parallelism. The quantity \(k_i\) isolates the reduction dimension and serves as a proxy for reuse and arithmetic intensity. Finally, \(d_i\) measures the imbalance of the equivalent \(M\times N\) contraction and acts as a compact descriptor of contraction geometry, which is relevant to mapping quality and memory-access regularity.

\subsection{Plan-level feature groups}

A complete contraction plan \(\pi\) is a variable-length sequence of \(S\) pairwise steps and must therefore be mapped to a fixed-size vector before it can be used in a tabular learning model. We do this through a set of plan-level aggregations that preserve the main structural aspects of the path. To emphasize the dominant region of the plan, we define the bottleneck exponent
\begin{equation}
c_{\max} = \max_i c_i,
\end{equation}
and the FLOP-weighted coefficients
\begin{equation}
w_i = \frac{2^{c_i-c_{\max}}}{\sum_{\ell=1}^{S} 2^{c_\ell-c_{\max}}}.
\end{equation}

The resulting aggregated features are organized into five interpretable groups. The \emph{complexity / critical-load} block captures the bottleneck, total work, near-critical tail, and overall heterogeneity of the plan. The \emph{sequence / granularity} block captures fragmentation and the presence of long tails of very small contractions. The \emph{parallelism dynamics} block measures how much output-side parallelism is available globally and at the bottleneck. The \emph{reduction / intensity} block characterizes whether the expensive part of the plan is reduction-rich and therefore more likely to benefit from reuse. Finally, the \emph{geometry / memory access} block summarizes how balanced or skewed the expensive contractions are, which is relevant to kernel efficiency and memory behavior.

\subsection{Final inventory and design principles}

The final feature space therefore combines a small set of local primitives with a deliberately rich set of plan-level aggregations. This partial redundancy is intentional: several aggregated descriptors are correlated, but they encode different hypotheses about what limits runtime in practice, such as a single catastrophic step, a broad near-critical tail, insufficient output parallelism in the dominant region, low reduction-driven reuse, unfavorable contraction geometry, or an accumulation of fine-grained overheads. Keeping these hypotheses explicit at the feature-design stage is useful because it allows subsequent ablation and feature-selection stages to determine which structural mechanisms carry the most predictive signal.

A key design choice is that the feature vector is derived entirely from the contraction plan. Consequently, the same plan yields the same feature representation regardless of the GPU on which it is later executed. What changes across backends is the measured runtime, and therefore the supervision signal used for learning and evaluation.

Table~\ref{tab:plan_features} summarizes the final fixed-size representation used before feature selection. Further technical details of the cost model and the feature blocks are provided in Appendix~\ref{app:feature_details}.

\begin{table*}[t]
\centering
\small
\caption{Plan-level features used to represent a contraction plan $\pi$ as a fixed-size vector. Feature groups are separated by horizontal rules. Features marked $\dagger$ are discarded after selection. $\flat$ and $\sharp$ are exclusive to $\mathrm{opt\_pair}$ and $\mathrm{opt\_ndcg}$, respectively; all other features are used by both optimized models.} 

\label{tab:plan_features}
\begin{tabular}{l l l p{0.33\textwidth}}
\toprule
Feature name & Definition & Block & Interpretation \\
\midrule
\texttt{$\mathrm{max\_cost}^\dagger$} & $\max_i c_i$ & Complexity / critical load & Bottleneck work exponent. \\
\texttt{$\mathrm{log2\_sum\_flops}^\dagger$} & $\log_2 \sum_i 2^{c_i}$ & Complexity / critical load & Accumulated work over the full plan. \\
\texttt{$\mathrm{topq\_mean\_cost}^\dagger$} & $|I_q|^{-1}\sum_{i\in I_q} c_i$ & Complexity / critical load & Size of the near-critical tail. \\
\texttt{$\mathrm{avg\_cost}^\dagger$} & $S^{-1}\sum_i c_i$ & Complexity / critical load & Mean work level. \\
\texttt{$\mathrm{std\_cost}^\sharp$} & $\operatorname{std}(c_i)$ & Complexity / critical load & Heterogeneity of the cost profile. \\
\midrule
\texttt{n\_steps} & $S$ & Sequence / granularity & Fragmentation of the plan and potential management overhead. \\
\texttt{frac\_tiny\_steps} & $S^{-1}\sum_i \mathbf{1}_{\{c_i \le c_{\max}-\tau\}}$ & Sequence / granularity & Fraction of steps far below the bottleneck scale. \\
\midrule
\texttt{max\_out\_rank} & $\max_i p_i$ & Parallelism dynamics & Peak output size; proxy for peak local intermediate size and parallel exposure. \\
\texttt{avg\_out\_rank} & $S^{-1}\sum_i p_i$ & Parallelism dynamics & Average output-size level across the plan. \\
\texttt{costw\_out\_rank} & $\sum_i w_i p_i$ & Parallelism dynamics & Output parallelism in the expensive part of the plan. \\
\texttt{p\_at\_max\_cost} & $|I_{\max}|^{-1}\sum_{i\in I_{\max}} p_i$ & Parallelism dynamics & Parallel exposure at the bottleneck. \\
\midrule
\texttt{max\_red\_rank} & $\max_i k_i$ & Reduction / intensity & Largest reduction dimension encountered. \\
\texttt{$\mathrm{costw\_red\_rank}^\flat$} & $\sum_i w_i k_i$ & Reduction / intensity & Reduction-richness of the expensive region. \\
\texttt{k\_at\_max\_cost} & $|I_{\max}|^{-1}\sum_{i\in I_{\max}} k_i$ & Reduction / intensity & Arithmetic-intensity proxy at the bottleneck. \\
\midrule
\texttt{max\_asym} & $\max_i d_i$ & Geometry / memory access & Worst contraction imbalance in the plan. \\
\texttt{avg\_asym} & $S^{-1}\sum_i d_i$ & Geometry / memory access & Global tendency toward skinny or square contractions. \\
\texttt{costw\_asym} & $\sum_i w_i d_i$ & Geometry / memory access & Shape imbalance of the expensive region. \\
\texttt{d\_at\_max\_cost} & $|I_{\max}|^{-1}\sum_{i\in I_{\max}} d_i$ & Geometry / memory access & Mapping quality at the bottleneck. \\
\bottomrule
\end{tabular}
\end{table*}

\section{Methodology}
\label{sec:meth}

This section defines the common methodology used by the two experimental studies that follow. It describes how the candidate plans and relevance labels are constructed, how the data are partitioned and evaluated without group leakage, how the feature set and ranking models are selected, and which hardware and software platforms are used. Sections~\ref{sec:exper} and~\ref{sec:cross} then apply this common setup to source-backend evaluation and backend-shift evaluation, respectively.

\subsection{Dataset Construction and Relevance Labels}
\label{ssec:data_labels}

Execution time is backend-dependent, as it varies with the GPU and software stack, whereas the model input is backend-independent because it is computed solely from the contraction plan. Backend dependence therefore enters through the target variable.

Let $T(P_i)$ denote the execution time of a candidate plan $P_i$ on the target GPU, and let $P_{\mathrm{mf}}$ be the corresponding MinFill plan for the same tensor network. We define the relative speedup
\begin{equation}
Sp_i = \frac{T(P_{\mathrm{mf}})}{T(P_i)} ,
\end{equation}
and the relevance label
\begin{equation}
y_i = \log(1 + Sp_i).
\end{equation}
Since ranking is defined within each circuit-specific group of plans, this target is aligned with the goal of selecting the fastest candidate while being less sensitive to global timing scale than absolute runtime. The logarithm reduces the dynamic range of speedups, preserves their ordering, and yields non-negative relevance values suitable for NDCG-based objectives.

We use MinFill as a baseline because it is a classical and widely adopted heuristic for elimination-order problems closely related to treewidth, which in turn is a key proxy for tensor-network contraction complexity~\cite{MaS08, GrK21}. In practice, it provides a simple, inexpensive, and reproducible reference that requires neither training data nor hyperparameter tuning, making it a natural point of comparison for learned ranking models. Moreover, MinFill and related heuristics are commonly used as baselines in the literature on treewidth and sparse elimination~\cite{Heg06, BlP93}, so outperforming it offers a clear and interpretable measure of the added value of our LTR approach.
    
To expose the model to a broad range of contraction behaviors, we built the dataset from a diverse set of quantum circuits, including both well-known benchmarks, such as MQTBench~\cite{QBW23b}, and more irregular instances generated using QXTools. The goal was to induce as much variation as possible in the contraction plans produced by different heuristics and in their measured execution on the GPU. Representative families include QFT-based circuits, Greenberger–Horne–Zeilinger (GHZ), variational quantum eigensolver (VQE), and several classes of random circuits, including random quantum circuit (RQC), among many others.

For each circuit, we generated seven candidate contraction plans using structurally diverse heuristics. MinFill and FlowCutter were obtained through QXTools; the remaining candidates were produced using a Girvan--Newman community-detection heuristic~\cite{GiN02}, the Jdrasil heuristic solver, the Tamaki local-improvement heuristic, and two MinFill-based heuristics submitted by Terrioux et al.\ to the Parameterized Algorithms and Computational Experiments (PACE) 2017 treewidth challenge~\cite{DKT18}. Each plan was then executed on the target GPU to measure its wall-clock time \(T(P_i)\).

For each contraction plan, one warm-up run was followed by 50 measured repetitions. The reported runtime is the mean across these repetitions. GPU synchronization was enforced immediately before and after each timed contraction-plan execution using \texttt{CUDA.synchronize()}. All computations were performed in single-precision (\texttt{Float32}) arithmetic. Plans, caches, and workspace memory were reused across repetitions to amortize setup overheads. Any out-of-memory or timeout event (10 minutes) caused the run to be aborted and marked as failed, and T
those plans were excluded from the subsequent analyses. Consequently, some circuit groups contain fewer than seven valid candidates. CPU–GPU transfers, memory allocation, plan creation, and just-in-time (JIT) compilation times were excluded from the measurement, as they are not part of the contraction itself and would introduce backend-dependent overheads that are not representative of the execution cost in the GPU.

This protocol was applied consistently to all contraction plans on
both GPU backends.

\subsection{Data Partitioning and Evaluation Protocol}
\label{ssec:partition_protocol}
The 225 circuit-specific ranking groups were partitioned into three disjoint subsets: a development set containing 138 groups, a locked in-distribution test set containing 25 groups, and an out-of-distribution set containing the 62 groups from the held-out QFT-based family. The development set was used for feature selection, cross-validation, hyperparameter optimization, learning-curve analysis, and final model fitting. The ID and OOD sets remained locked throughout model development. All valid candidate plans associated with the same circuit instance were assigned to the same subset, thereby preventing circuit-level data leakage. The resulting partition is summarized in Table~\ref{tab:data_split}.

\begin{table*}[t]
\small
\centering
\begin{tabular}{p{2.5cm}p{6cm}p{4cm}}
\toprule
\textbf{Block} & \textbf{Content} & \textbf{Allowed Use} \\
\midrule

Development & 138 circuit groups & Feature selection, grouped cross-validation, hyperparameter optimization, learning curves, and final model fitting \\
Test (ID) & 25 unseen circuit groups from the same domain & Final evaluation \\
Out-of-distribution (OOD) & 62 groups from the held-out QFT-based family & Circuit-family-shift evaluation \\
\bottomrule
\end{tabular}
\caption{Partition of the 225 circuit groups used for model development and final evaluation.}
\label{tab:data_split}
\end{table*}

\subsection{Feature Validation and Selection}
\label{ssec:feature_validation}

The initial feature representation is first tested with a diagnostic ranking model and group-aware validation. Sanity checks and learning curves are then used to verify that the observed signal is not caused by leakage or trivial artifacts before redundant features are removed. We trained the ranking models with XGBoost.jl~\cite{ChG16}, using either \texttt{rank:ndcg} or \texttt{rank:pairwise} as the objective.
Plans were grouped by circuit so that comparisons were restricted to candidates belonging to the same ranking problem.
    
To evaluate the raw predictive value of the initial 18-feature
representation, we began with a simple heuristic configuration. We selected a set of ``rule-of-thumb'' hyperparameters to evaluate the raw predictive power of the 18-feature vector $(\mathbf{x}_{18})$. These parameters were chosen based on established literature for gradient boosting on medium-sized datasets~\cite{ChG16, MuG16} to ensure a balance between learning capacity and training speed:

\begin{itemize}
    \item \textbf{Objective (\texttt{rank:ndcg} or \texttt{rank:pairwise}): } Selected to align the model with the LTR paradigm.
    \item \textbf{Learning Rate (\texttt{eta} $= 0.01$):} A small learning rate was used to introduce gradual updates and reduce the risk of overfitting.
    \item \textbf{Tree Depth (\texttt{max\_depth} $= 6$):} A moderate depth used to capture non-linear interactions between computational cost and tensor shapes without inducing immediate overfitting.
    \item \textbf{Tree Method (\texttt{gpu\_hist}):} Employed to leverage the computational throughput of the GPU during the histogram construction of the trees~\cite{MFH18}.
    \item \textbf{Regularization ($\lambda = 1, \alpha = 0$):} Standard L2 regularization to stabilize the initial weight updates.
    \item \textbf{Subsampling (\texttt{subsample}=0.7):} A fraction of the training instances was sampled for each tree to reduce overfitting and improve robustness.
\end{itemize}

We evaluated the diagnostic models using five-fold grouped cross-validation over the complete development set of 138 circuit groups. In each iteration, four folds were used for training and the remaining fold for validation. All candidate plans associated with the same circuit were assigned to the same fold, thereby preventing circuit-level leakage and preserving the grouped ranking structure. Validation and early stopping were based on NDCG@3 computed on the validation fold of the corresponding iteration. The normalized discounted cumulative gain truncated at the top three positions (NDCG@3) measures how well the most relevant plans are placed near the top of each group. Feature selection and the learning curves were performed using five-fold grouped cross validation over the complete development set. 

We then subjected this initial model to standard sanity checks. In particular, we verified that performance collapsed when the labels were randomly shuffled, confirming that the predictive signal did not arise from leakage or trivial dataset artifacts. After this preliminary validation, we computed learning curves to assess the quality and generalization behavior of the initial model. These curves revealed clear signs of overfitting, with training performance continuing to improve while validation performance saturated or degraded. This diagnosis motivated the next stages of the protocol, aimed at improving robustness and reducing unnecessary model complexity.

Feature selection was guided by two complementary importance analyses: Gain, which reflects the reduction in loss contributed by a feature during tree splits, and Shapley values, which provide a more global view of feature contribution~\cite{GDB00, LuL17}. This analysis showed that 4 of the original 18 features were either highly redundant or contributed negligible predictive signal, and they were therefore removed. The union of the two selected feature sets contained 14 features. Each optimized model used 13 features: 12 shared features and one objective-specific feature, as indicated in Table~\ref{tab:plan_features}.

\subsection{Model Optimization and Training}
\label{ssec:model_training}

This stage selects the final hyperparameters, trains the two optimized rankers, and defines the comparison baselines used in the source-backend evaluation.

We optimized the XGBoost ranking models through a randomized hyperparameter search using the same five-fold grouped cross-validation protocol. The search focused on a small set of parameters with the greatest impact on model capacity and regularization: \texttt{max\_depth}, which controls tree complexity;
\texttt{eta}, which sets the learning rate; and \texttt{lambda}, \texttt{alpha}, and \texttt{min\_child\_weight}, which regulate model smoothness and splitting conservativeness. These parameters were explored for both the \texttt{rank:ndcg} and \texttt{rank:pairwise} objectives.

Each sampled configuration was evaluated using the mean validation \(\mathrm{NDCG}@3\) across the five folds, with early stopping applied independently on the validation fold of each iteration to determine
the appropriate number of boosting rounds. The best-performing configurations were retained as the final optimized models, whose main hyperparameters are summarized in Table~\ref{tab:opt_models}.

Our main comparison focuses on two optimized XGBoost rankers that differ only in the training objective. The first, \texttt{opt\_ndcg}, uses \texttt{rank:ndcg}, whereas the second, \texttt{opt\_pair}, uses \texttt{rank:pairwise}. 
Both models were obtained with the same protocol described above: the same data partitions, the same feature-selection procedure, the same hyperparameter-optimization pipeline, and the same validation metric, \(\mathrm{NDCG}@3\). The comparison therefore isolates the effect of the training objective while keeping the rest of the pipeline fixed.

This distinction is meaningful because the two objectives embody different ranking strategies. The \texttt{rank:ndcg} objective uses a LambdaMART formulation scaled toward NDCG, making it explicitly sensitive to the quality of the top part of the ranking, whereas \texttt{rank:pairwise} uses the original pairwise RankNet-style loss and focuses on reducing pairwise inversions. XGBoost also provides a third built-in ranking objective, \texttt{rank:map}, but this objective is primarily intended for binary relevance labels. Since our target is graded rather than binary, \texttt{rank:ndcg} and \texttt{rank:pairwise} are the two built-in objectives most appropriate for the present study. XGBoost does not provide a dedicated pointwise objective for grouped LTR, so the comparison is centered on the two ranking objectives that best match our task.

\begin{table*}[t]
\centering
\small
\caption{Main hyperparameters of the two optimized XGBoost ranking models. A dash indicates that the parameter was not explicitly fixed and was left at the library default. L1 and L2 denote two regularization techniques to control overfitting.}
\label{tab:opt_models}
\begin{tabular}{llll}
\toprule
\textbf{Parameter} & \textbf{\texttt{opt\_ndcg}} & \textbf{\texttt{opt\_pair}} & \textbf{Description} \\
\midrule
\texttt{objective} & \texttt{rank:ndcg} & \texttt{rank:pairwise} & Training objective \\
\texttt{eval\_metric} & \texttt{ndcg@3} & \texttt{ndcg@3} & Validation metric \\
\texttt{maximize} & true & true & Maximize evaluation metric \\
\texttt{eta} & 0.01 & 0.05 & Learning rate \\
\texttt{max\_depth} & 6 & 5 & Maximum tree depth \\
\texttt{subsample} & 0.7 & 0.5 & Fraction of samples used per tree \\
\texttt{colsample\_bytree} & -- & 0.5 & Fraction of features sampled per tree \\
\texttt{min\_child\_weight} & -- & 3 & Minimum child-node weight \\
\texttt{gamma} & -- & 0.5 & Minimum loss reduction for a split \\
\texttt{lambda} & -- & 0.0 & L2 regularization \\
\texttt{alpha} & -- & -- & L1 regularization  \\
\texttt{seed} & 42 & 42 & Random seed \\
\bottomrule
\end{tabular}
\end{table*}

To contextualize the performance of the learned rankers, we compare them against three baselines. The first, \texttt{random}, provides a lower bound by assigning the plan order uniformly at random. The second, \texttt{mf\_ndcg}, is a hybrid baseline that forces the MinFill plan to the first position and uses \texttt{opt\_ndcg} only to rank the remaining candidates. The third, \texttt{mf\_rand}, also fixes MinFill at the first position but assigns the remaining plans randomly.

These two MinFill-based baselines are useful because the target speedup is defined relative to the MinFill plan. As a result, they allow us to separate two effects: first, the value of allowing the learned model to challenge MinFill at the top of the ranking (by comparing \texttt{opt\_ndcg} against \texttt{mf\_ndcg}); and second, the value of learning a meaningful secondary ranking once MinFill is fixed at Top-1 (by comparing \texttt{mf\_ndcg} against \texttt{mf\_rand}). Together, these comparisons clarify whether the gains of the learned models come mainly from improving the first recommendation, from improving the ordering of the remaining plans, or from both.

After feature selection and hyperparameter optimization (HPO), the selected model configurations were retrained on the full development set,i.e., all 138 groups.The resulting fixed models were then evaluated once on the locked ID and OOD sets. Their robustness and generalization are assessed experimentally through the source-backend and backend-shift evaluations presented in Sections~\ref{sec:exper} and~\ref{sec:cross}.

After completing the full experimental evaluation on the locked in-distribution and OOD splits, a separate deployment model may be retrained using all labeled data to maximize practical utility. This post-evaluation refit is intended for downstream use only and should not be interpreted with the same reported test/OOD metrics, which apply to the evaluation protocol described above.

\subsection{Experimental Platform and Software}
\label{ssec:hardware}

The following platform is shared by the experimental analyses in Sections~\ref{sec:exper} and~\ref{sec:cross}. The RTX A6000 provides the source-backend measurements used for model development and the main evaluation, while the Tesla V100 is reserved for the backend-shift study.

The primary execution environment was a high-performance computing server with two AMD EPYC 7282 processors, each providing 16 cores at a base clock frequency of 2.8\,GHz, together with 256\,GiB of DDR4 memory and 64\,MiB of L3 cache.

The NVIDIA RTX A6000 was used as the primary GPU backend to generate the labels employed during model development and to conduct all learning-curve, in-distribution, and out-of-distribution experiments reported in Section~\ref{sec:exper}. The NVIDIA Tesla V100 was used only for the cross-GPU analyses presented in Section~\ref{sec:cross}. Measurements from the complete dataset were used to compare the GPU-specific ground-truth rankings, whereas the locked in-distribution test subset was used to evaluate the zero-shot transfer of the Ampere-trained models. The main characteristics of both GPU backends are summarized in Table~\ref{tab:GPU_Architectures}.

\begin{table*}[t]
\small
\centering
\caption{GPU specifications used in the experimental study.}
\label{tab:GPU_Architectures}
\begin{tabular}{@{}lll@{}}
\toprule
\textbf{Feature} & 
\shortstack[l]{\textbf{NVIDIA}\\\textbf{RTX A6000}} & 
\shortstack[l]{\textbf{NVIDIA Tesla V100}\\\textbf{PCIe 32GB}} \\
\midrule
Architecture           & Ampere        & Volta        \\
Release Date           & October 2020  & March 2018   \\
CUDA Cores             & 10,752        & 5,120        \\
Tensor Cores           & 336 (3rd Gen) & 640 (1st Gen)\\
RT Cores               & 84 (2nd Gen)  & None         \\
Memory                 & 48 GB GDDR6   & 32 GB HBM2   \\
Memory Bandwidth       & 768.0 GB/s    & 897.0 GB/s   \\
FP32 Performance       & 38.71 TFLOPS  & 14.13 TFLOPS \\
Interface              & PCIe 4.0 x16  & PCIe 3.0 x16 \\
Max Power Consumption  & 300 W         & 250 W        \\
\bottomrule
\end{tabular}
\end{table*}

The simulations were implemented in Julia using QXTools.jl to construct the tensor networks and candidate contraction plans~\cite{BOH22}, OMEinsum.jl to execute the contractions, and CUDA-enabled GPU execution. The ranking models were trained with XGBoost.jl~\cite{ChG16,XGBoostJL253}. The same software stack was used
on both GPU backends. Complete package versions, environment manifests, and system-configuration details will be provided in the accompanying repository.

\section{Source-Backend Evaluation}
\label{sec:exper}

This section evaluates the learning behavior and final decision quality of the optimized rankers on the RTX A6000, the source backend used to generate their training labels. We first use learning curves to examine sample efficiency and train--validation behavior within the development data. We then evaluate the fixed models on the locked in-distribution test set and on the held-out QFT-based family, separating standard generalization from circuit-family shift.

\subsection{Learning-Curve Analysis}
\label{ssec:learning_curves}

To assess the sample efficiency and generalization behavior of the two optimized models, we analyze learning curves built exclusively on the development set. We report two complementary metrics. The first is \(\mathrm{NDCG}@3\), which is the validation metric used throughout model selection and therefore provides the most direct view of how ranking quality evolves as additional training circuits are incorporated. The second is \(\mathrm{Regret3}\), a decision-oriented metric that measures the relative speedup loss incurred when the best plan within the model's Top-3 recommendations is selected. While \(\mathrm{NDCG}@3\) evaluates the quality of the ranking itself, \(\mathrm{Regret3}\) captures the practical cost of imperfect ranking decisions. For both metrics, we report values on the training subset and on a held-out validation subset.

\begin{figure*}[t]
    \centering
    \begin{subfigure}[b]{0.48\textwidth}
        \centering
        \includegraphics[
            width=\textwidth,
            page=1,
            trim={0 25mm 0 0},
            clip
        ]{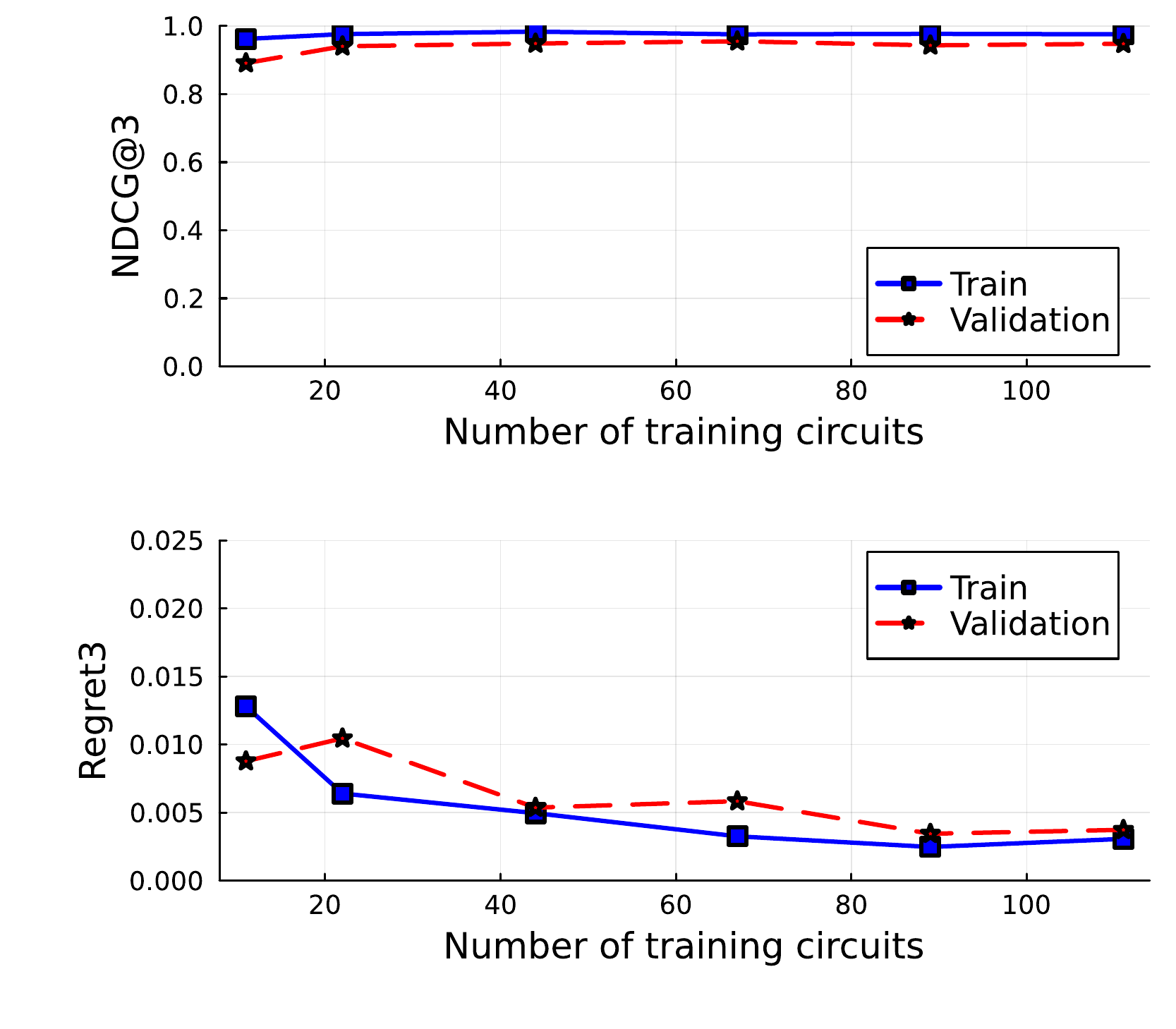}
        \caption{\texttt{opt\_ndcg}}
        \label{fig:ndcg3_curves}
    \end{subfigure}
    \hfill
    \begin{subfigure}[b]{0.48\textwidth}
        \centering
        \includegraphics[
            width=\textwidth,
            page=1,
            trim={0 25mm 0 0},
            clip
        ]{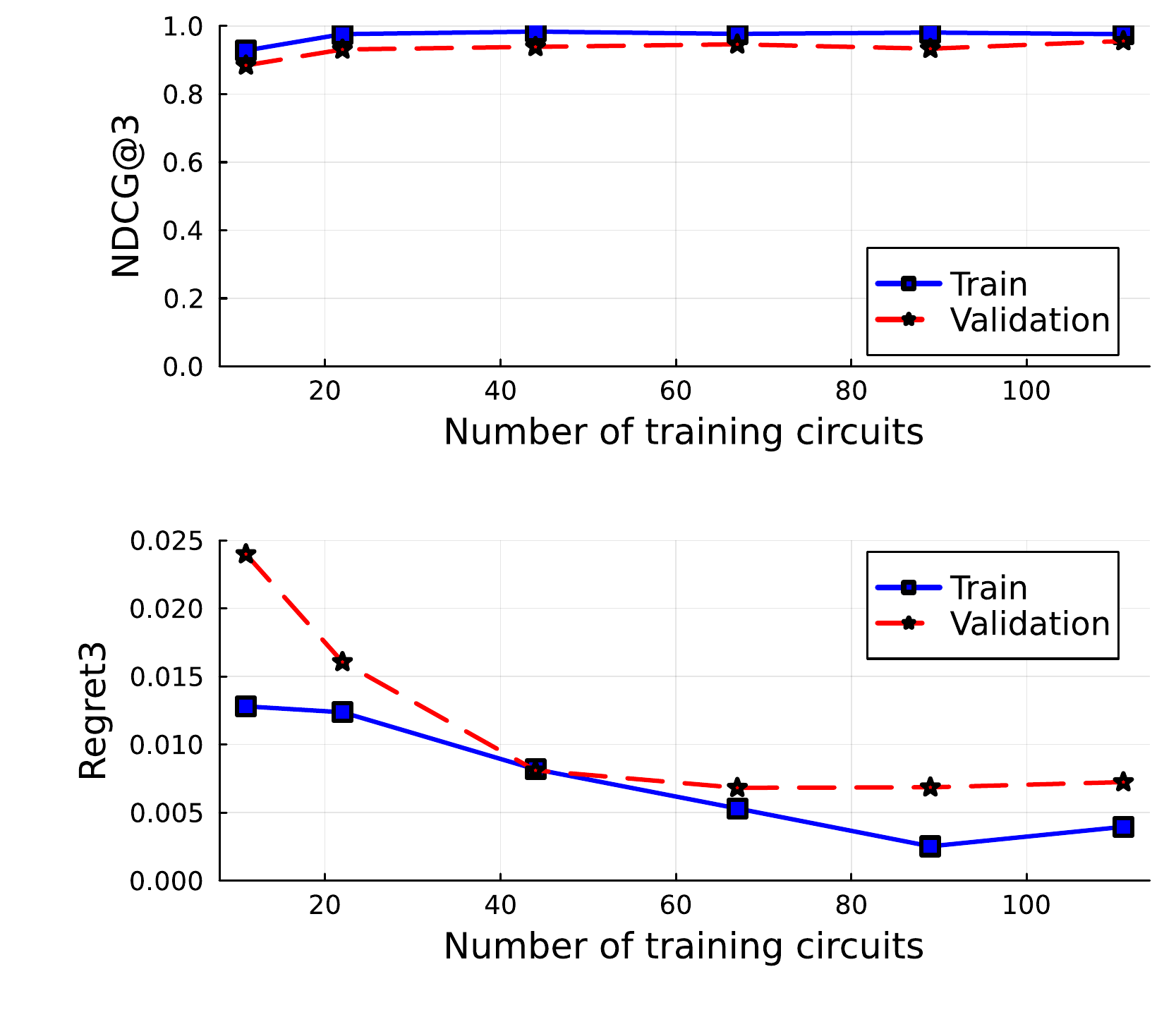}
        \caption{\texttt{opt\_pair}}
        \label{fig:pairwise_curves}
    \end{subfigure}

    \caption{Learning curves for the two optimized models. Each subfigure reports training and validation \(\mathrm{NDCG}@3\) (top) and \(\mathrm{Regret3}\) (bottom) as a function of the number of training circuit groups.}
    \label{fig:learning_curves}
\end{figure*}

Fig.~\ref{fig:learning_curves} shows that both models maintain high \(\mathrm{NDCG}@3\) values throughout the experiment and low \(\mathrm{Regret3}\), with limited train--validation gaps after the final optimization stages. The most visible improvements occur up to roughly 40--45 training circuits, after which the curves enter a broad plateau. This behavior suggests that the selected features are already informative at moderate sample sizes and that additional data yields diminishing, although still positive, returns.

The same figure also indicates that the strong overfitting observed in the earlier diagnostic models has been substantially reduced. In both optimized models, the training and validation curves remain reasonably close across the full range of dataset sizes. It would nevertheless be too strong to claim a complete absence of overfitting: a non-negligible train--validation gap remains, especially at smaller training sizes, but it stays limited and does not widen as more data are added.

In terms of \(\mathrm{NDCG}@3\), both \texttt{opt\_ndcg} and \texttt{opt\_pair} achieve high validation values, indicating strong ranking quality even with relatively small development subsets. Because the curves remain close to the upper bound of the metric, the absolute gains are necessarily modest and should be interpreted as refinements near the performance ceiling rather than as large raw improvements. Within this regime, \texttt{opt\_ndcg} follows a slightly more regular validation trajectory and tends to maintain a somewhat smaller train--validation gap. This is consistent with the fact that its training objective is directly aligned with the validation metric.

The distinction becomes clearer in \(\mathrm{Regret3}\). For both models, regret decreases noticeably as more training circuits are added, confirming that the learned rankings become increasingly useful for practical Top-3 selection. However, \texttt{opt\_ndcg} reaches consistently lower validation regret in the medium- and large-data regime, ending near \(3\times 10^{-3}\) to \(4\times 10^{-3}\), whereas \texttt{opt\_pair} stabilizes around \(7\times 10^{-3}\). Thus, although \texttt{opt\_pair} remains competitive as a ranking model, \texttt{opt\_ndcg} yields better decision quality when the final goal is to minimize the speedup loss within the Top-3 recommendations.

The learning-curve analysis supports two conclusions. First, both optimized models are sample-efficient and generalize reasonably well once feature selection and hyperparameter optimization have been applied. Second, the NDCG-aware objective provides a better trade-off between ranking quality and downstream decision quality, which anticipates the stronger performance of \texttt{opt\_ndcg} in the final test-set comparisons.

\subsection{In-Distribution and Circuit-Family-Shift Evaluation}
\label{ssec:id_ood}

We next compare the two optimized models, \texttt{opt\_ndcg} and \texttt{opt\_pair}, against the three baselines introduced above: \texttt{random}, \texttt{mf\_ndcg}, and \texttt{mf\_rand}. We first evaluate them on the standard held-out ID test split and then on a more challenging OOD domain-shift split. This two-stage evaluation allows us to assess both the absolute quality of the learned rankings under standard conditions and their robustness when the test circuits differ more substantially from those seen during development.

To characterize model performance, we report both hit-based and decision-oriented metrics. Top-1 and Top-3 denote the percentage of groups in which the true best plan is ranked first or appears within the first three positions, respectively. To quantify the practical cost of imperfect recommendations, we also report \(\mathrm{Regret1}\) and \(\mathrm{Regret3}\), defined as
\begin{align}
\mathrm{Regret1} &= 1-\frac{Sp_1}{Sp^\star},\\
\mathrm{Regret3} &= 1-\frac{\max_{i\in \mathrm{Top-3}} Sp_i}{Sp^\star},
\end{align}
where \(Sp^\star\) is the optimal speedup in the group and \(Sp_1\) is the speedup of the plan at rank 1. Lower regret therefore indicates better decision quality. Finally, we report \emph{win rate}, defined as the fraction of groups in which the model's top-ranked plan achieves a higher speedup than the MinFill plan, with ties counted as half a win for each side. Since \texttt{mf\_ndcg} and \texttt{mf\_rand} always place MinFill at rank 1, their win rate is \(50\%\) by construction.

\subsubsection{In-distribution test}

\begin{table*}[t]
\small
\centering
\caption{ID test results (N = 25 circuit groups). Higher is better for Top-1, Top-3, and \(\mathrm{Win\ rate}\); lower is better for \(\mathrm{Regret1}\) and \(\mathrm{Regret3}\).}
\label{tab:test_results}
\begin{tabular}{lccccc}
\toprule
Model & Top-1 (\%) & Top-3 (\%) & Regret1 & Regret3 & Win rate \\
\midrule
\texttt{opt\_ndcg} & \textbf{60.00} & \textbf{96.00} & \textbf{0.0141} & \textbf{0.0004} & \textbf{92.00} \\
\texttt{opt\_pair}          & \textbf{60.00} & 92.00          & 0.0211          & \textbf{0.0004} & 88.00 \\
\texttt{random}    & 22.95          & 58.44          & 0.0699          & 0.0137          & 70.16 \\
\texttt{mf\_ndcg}  & 12.00          & 84.00          & 0.0735          & 0.0025          & 50.00 \\
\texttt{mf\_rand}  & 12.00          & 28.00          & 0.0735          & 0.0319          & 50.00 \\
\bottomrule
\end{tabular}
\end{table*}

Table~\ref{tab:test_results} reveals a clear hierarchy on the ID test split. The two optimized models substantially outperform all baselines across the main decision metrics, confirming that the selected structural features capture non-trivial patterns that generalize to unseen circuits from the same domain. The comparison between the two optimized models is also consistent with the learning-curve analysis: both achieve the same Top-1 accuracy (\(60\%\)), but \texttt{opt\_ndcg} attains better Top-3 (\(96\%\) vs.\ \(92\%\)), lower \(\mathrm{Regret1}\) (\(0.0141\) vs.\ \(0.0211\)), and the same best \(\mathrm{Regret3}\) (\(0.0004\)). Thus, both models usually place an excellent candidate near the top of the ranking, but \texttt{opt\_ndcg} is more reliable when only a single plan will ultimately be executed.

The baselines help clarify where these gains come from. The \texttt{random} baseline performs much worse than either optimized model in Top-1, Top-3, and regret, showing that the observed gains are not due to chance. The two MinFill-based baselines are more informative still. Because they force MinFill to rank first, both inherit very poor Top-1 and \(\mathrm{Regret1}\), which shows that MinFill is rarely the true optimum in this setting. However, comparing \texttt{mf\_ndcg} against \texttt{mf\_rand} isolates the value of ordering the remaining candidates well once MinFill is fixed at the top: Top-3 rises from \(28\%\) to \(84\%\), and \(\mathrm{Regret3}\) drops from \(0.0319\) to \(0.0025\). This confirms that the secondary ranking still matters greatly when the user is willing to inspect a shortlist, although it cannot compensate for a systematically poor first recommendation.

\subsubsection{Circuit-family-shift evaluation}

\begin{table*}[t]
\small
\centering
\caption{OOD domain-shift results (N = 62 circuit groups). Higher is better for Top-1, Top-3, and \(\mathrm{Win\ rate}\); lower is better for \(\mathrm{Regret1}\) and \(\mathrm{Regret3}\).}
\label{tab:ood_results}
\begin{tabular}{lccccc}
\toprule
Model & Top-1 (\%) & Top-3 (\%) & Regret1 & Regret3 & Win rate \\
\midrule
\texttt{opt\_ndcg} & \textbf{38.71} & \textbf{62.90} & \textbf{0.0520} & \textbf{0.0181} & 73.39 \\
\texttt{opt\_pair}          & 30.65          & \textbf{62.90} & 0.0800          & 0.0234          & \textbf{87.10} \\
\textit{random}    & 27.87          & 54.65          & 0.0775          & 0.0265          & 70.97 \\
\texttt{mf\_ndcg}  & 8.06           & 61.29          & 0.0784          & 0.0189          & 50.00 \\
\texttt{mf\_rand}  & 8.06           & 43.55          & 0.0784          & 0.0273          & 50.00 \\
\bottomrule
\end{tabular}
\end{table*}

For the OOD evaluation, we constructed a dedicated held-out split by reserving all QFT-based circuits, across the considered qubit counts, and excluding them entirely from training, validation, and the in-distribution test set. The OOD results in Table~\ref{tab:ood_results} therefore measure the ability of the models to transfer to a circuit family not seen at any stage of model development.

As expected, all models degrade under this domain shift. The learned
rankers retain useful ranking information, but their advantage over
the baselines is no longer uniform across all decision metrics.
\texttt{opt\_ndcg} provides the strongest OOD performance, achieving
the best Top-1, Regret1, and Regret3 values while tying
\texttt{opt\_pair} on Top-3. This indicates that it remains the most effective option when the goal is to get as close as possible to the true optimum on an unseen circuit family.

Taken together, Tables~\ref{tab:test_results} and~\ref{tab:ood_results} support three conclusions. First, both learned models clearly outperform the random and MinFill-based baselines in the standard ID setting. Second, \texttt{opt\_ndcg} provides the strongest overall trade-off between ranking quality and decision quality, especially when a single recommendation must be trusted. Third, the OOD experiment confirms that transfer performance is sensitive to the held-out circuit family: although the learned rankers retain useful predictive signal, the degradation from in-distribution to OOD indicates that generalization quality depends noticeably on which family is excluded during development. In our case, holding out QFT-based circuits provides a demanding but still single-family domain-shift test, so these results should be interpreted as evidence of non-trivial sensitivity to unseen circuit structure rather than as a definitive limitation of the feature set. This suggests that broader training distributions and additional family-level hold-out experiments would be valuable in future work.

\section{Backend-Shift Evaluation}
\label{sec:cross}

Section~\ref{sec:exper} evaluated the rankers on the RTX A6000, which also supplied the measurements used during model development. This section examines what changes when the same candidate plans are executed on the Tesla V100 and asks whether the Ampere-trained rankings remain useful without retraining or target-backend calibration.

The study separates two questions. First, using all circuit groups, we compare the empirical rankings obtained by executing the same candidate plans on both GPUs, which shows how much the ranking target itself changes across backends. Second, on the locked in-distribution test set, we keep the Ampere-trained models and their predicted orderings unchanged and evaluate those orderings against speedups and relevance labels recomputed from the Volta measurements. This isolates zero-shot model transfer from changes in the candidate set or test composition. The results describe one observed backend shift and should not be interpreted as evidence of general GPU portability.

\subsection{Ground-Truth Ranking Stability Across GPUs}
\label{ssec:ground}

Before evaluating model transfer, we first examine whether changing the GPU alters the ranking target itself. This analysis uses all 225 circuit groups in the dataset and does not involve model predictions. For each circuit, the same set of successfully measured candidate plans were executed on the RTX A6000 and the Tesla V100, and the two empirical rankings were obtained from the corresponding execution measurements.

We compare the complete rankings using Kendall's \(\tau_b\)~\cite{KeG90}, computed independently within each circuit group. This coefficient measures pairwise ordinal agreement and accounts for ties: a value of \(1\) corresponds to identical orderings, whereas lower values indicate that more pairwise preferences change between GPUs. 

To quantify the sampling variability of the aggregate cross-GPU stability measures, we computed 95\% percentile-bootstrap confidence intervals by resampling complete circuit groups. For each of \(B = 10{,}000\) bootstrap replicates, \(N = 225\) groups were sampled with replacement from the observed dataset, and the mean Kendall's \(\tau_b\), same-best-plan rate, and mean Top-3 overlap were recomputed. The interval endpoints were defined by the 2.5th and 97.5th percentiles of the resulting bootstrap distribution. Resampling was performed at the circuit-group level (\texttt{base\_id}), rather than at the level of individual contraction plans, to preserve the dependence among candidate plans retained for the same circuit. The random seed was fixed to 42 for reproducibility. These intervals quantify variability across circuit groups under the assumption that the observed dataset is representative of the broader population of circuits of interest; they do not account for timing-measurement uncertainty or for changes in circuit-family composition.

Table~\ref{tab:cross_gpu_tau_new} summarizes the distribution of the resulting coefficients.

\begin{table}[t]
    \centering
    \small
    \caption{Summary of Kendall's \(\tau_b\) across the 225 circuit groups.}
    \label{tab:cross_gpu_tau_new}
    \begin{tabular}{lcc}
        \toprule
        \textbf{Statistic} & \textbf{Value} & \textbf{95\% CI} \\
        \midrule
       
        Mean               & 0.845 & [0.808, 0.878] \\
        Median             & 1.000 & [1.000, 1.000] \\
        Standard deviation & 0.273 & ---          \\
        Minimum            & -0.143 & ---          \\
        Maximum            & 1.000 & ---          \\
        
        \bottomrule
    \end{tabular}
\end{table}

The upper part of the ranking is examined separately because it is the most relevant region for plan selection. Let \(I^{(3)}_{g,A}\) and \(I^{(3)}_{g,B}\) denote the unordered sets of the three fastest plans for circuit \(g\) on the two GPUs. The Top-3 set overlap for that circuit is

\begin{equation}
\mathrm{Ov}_{3,g}
=
\frac{\left|I^{(3)}_{g,A}\cap I^{(3)}_{g,B}\right|}{3},
\end{equation}

\noindent
and the value reported in Table~\ref{tab:cross_gpu_decision_new} is its mean across the \(G\) circuit groups,

\begin{equation}
\mathrm{Ov}_3
=
\frac{1}{G}
\sum_{g=1}^{G} \mathrm{Ov}_{3,g}.
\end{equation}

The order of the plans within each Top-3 set is ignored. Thus, a circuit contributes \(1\), \(2/3\), \(1/3\), or \(0\), depending on whether the two GPUs share three, two, one, or none of their three fastest plans. The same-best-plan rate is stricter at the first position and measures the fraction of circuits for which both GPUs identify the same fastest plan.

\begin{table}[t]
    \centering
    \small
    \caption{Stability of the decision-relevant part of the rankings across the 225 circuit groups.}
    \label{tab:cross_gpu_decision_new}
    \begin{tabular}{lcc}
        \toprule
        \textbf{Metric} & \textbf{Mean} & \textbf{95\% CI} \\
        \midrule
        Same-best-plan rate       & 84.0\% & [79.1\%, 88.4\%] \\
        Mean Top-3 set overlap    & 89.63\% & [87.1\%, 92.0\%] \\
        \bottomrule
    \end{tabular}
\end{table}

The results in Tables~\ref{tab:cross_gpu_tau_new} and~\ref{tab:cross_gpu_decision_new} indicate that most pairwise preferences are preserved when moving from Ampere to Volta, although the degree of stability varies among circuits. The mean \(\tau_b\) is \(0.85\), while the median of \(1.0\) means that at least half of the circuit groups retain the complete ordering of their common candidate plans. The lower values found in some groups show that this stability is not universal and that the backend can substantially reorder the candidates in particular cases.

Agreement is slightly weaker when only the fastest plan is considered. The two GPUs select the same best plan in \(84\%\) of the circuits, so local exchanges near the top of the ranking occur more often than complete changes in ranking structure. The mean Top-3 set overlap is \(89.63\%\), corresponding to an average of \(2.69\) shared plans out of three per circuit. The short list of competitive plans is therefore more stable than the identity of the single fastest plan, which is useful when several recommendations can be tested on the target GPU.

Stability also differs across circuit families, as reported in Table~\ref{tab:cross_gpu_families_new}. Quantum Walk circuits retain exactly the same rankings on both devices, while Grover and GHZ circuits also show high agreement. RQC circuits occupy an intermediate position, and the lower values for the QFT-based family indicate greater sensitivity to the backend.

\begin{table*}[t]
	\centering
	\small
    \caption{Ground-truth ranking stability by circuit family.}
    \label{tab:cross_gpu_families_new}
	\begin{tabular}{lcccc}
		\toprule
		\textbf{Family} & \textbf{N} & \textbf{Mean \(\tau_b\)} & \textbf{Same-best-plan rate} & \textbf{Top-3 overlap} \\
		\midrule

        GHZ & 18 & 0.958 & 94.4\% & 96.3\% \\
        Grover & 7 & 0.959 & 85.7\% & 95.2\% \\
        QNN & 24 & 0.879 & 87.5\% & 93.1\% \\
        Quantum Walk & 5 & 1.000 & 100.0\% & 100.0\% \\
        RQC & 86 & 0.874 & 88.4\% & 91.9\% \\
        VQE & 7 & 0.864 & 85.7\% & 95.2\% \\
        Deutsch--Jozsa & 3 & 0.778 & 100.0\% & 88.9\% \\
        Ground State & 1 & 1.000 & 100.0\% & 100.0\% \\
        Pricing Call & 11 & 0.931 & 90.9\% & 97.0\% \\
        Portfolio QAOA & 1 & 1.000 & 100.0\% & 100.0\% \\
        QFT-based & 62 & 0.710 & 69.4\% & 79.6\% \\

		\bottomrule
	\end{tabular}
\end{table*}

A plausible explanation is that the regular contraction structures produced by QWALK and GHZ circuits leave less room for hardware-dependent trade-offs to change the ordering. More heterogeneous plans may contain closer balances between computational work, reduction size, parallelism, contraction geometry, and memory traffic, allowing architectural differences to affect their relative execution times. The present experiment does not isolate these factors, so this interpretation should not be treated as a causal result.

This analysis shows how much the empirical ranking target changes between the two GPUs. It does not establish whether the Ampere-trained models still make good recommendations on Volta, which is evaluated separately in the next subsection.

\subsection{Zero-Shot Model Transfer under GPU Backend Shift}
\label{ssec:transfer}

We next examine whether the rankers trained from Ampere measurements remain useful when plan quality is determined by execution on Volta. This evaluation uses the same 25 circuit groups and the same candidate plans as the locked in-distribution test. The models are not retrained or calibrated for the target GPU. Since their input features depend only on the contraction plans, the feature vectors and predicted rankings remain unchanged; only the runtimes, MinFill-relative speedups, and relevance labels used to evaluate those predictions are recomputed from the Volta executions.

Table~\ref{tab:hardware_generalisation} compares the decision quality of both rankers when their fixed predictions are evaluated against the Ampere and Volta ground truths. Top-1, Top-3, Regret1, Regret3, and win rate are computed in the same way as in the earlier in-distribution evaluation, with MinFill executed separately on each backend.

\begin{table*}[t]
\small
\centering
\caption{Zero-shot cross-GPU performance on the locked in-distribution test set (\(N=25\)). Both models were trained using Ampere measurements. The evaluation backend determines the runtimes, speedups, and relevance labels used to assess the fixed model predictions.}
\label{tab:hardware_generalisation}
\begin{tabular}{lcccccc}
\toprule
\textbf{Model} & \textbf{Evaluation backend} & \textbf{Top-3 (\%)} & \textbf{Regret3} & \textbf{Top-1 (\%)} & \textbf{Regret1} & \textbf{Win rate (\%)} \\
\midrule
\texttt{opt\_ndcg} & Ampere & 96.0 & 0.0004 & 60.0 & 0.0141 & 92.0 \\
\texttt{opt\_ndcg} & Volta  & 92.0 & 0.0068 & 64.0 & 0.0254 & 92.0 \\
\addlinespace
\texttt{opt\_pair} & Ampere & 92.0 & 0.0004 & 60.0 & 0.0211 & 88.0 \\
\texttt{opt\_pair} & Volta  & 88.0 & 0.0125 & 52.0 & 0.0531 & 88.0 \\
\bottomrule
\end{tabular}
\end{table*}

Both models retain a high Top-3 rate on Volta, with a decrease of four percentage points in each case. The shortlists obtained from the structural plan features therefore remain useful after the backend change. The increase in regret reveals a loss that Top-3 alone does not capture: when the fastest plan is absent from the shortlist, the best available recommendation can be farther from the optimum on Volta.

The NDCG-oriented model is less affected in this experiment. Its Regret1 increases from \(0.014\) to \(0.025\), whereas the pairwise model rises from \(0.021\) to \(0.053\). Regret3 follows the same pattern. This suggests that \texttt{opt\_ndcg} preserves the quality of its highest-ranked candidates better under the observed backend shift. The comparison involves only one target architecture, however, and does not establish that the NDCG-oriented objective is generally more robust across GPUs.

The Top-1 rate of \texttt{opt\_ndcg} increases from \(60\%\) to \(64\%\), while that of \texttt{opt\_pair} decreases to \(52\%\). The increase for \texttt{opt\_ndcg} corresponds to one additional correct circuit among the 25 test groups and should not be interpreted as an improvement caused by the change of GPU. With a small test set, an exchange between two closely performing plans can change the Top-1 result for an entire circuit without representing a broad improvement in ranking quality.

The win rate remains unchanged for both models: \(92\%\) for \texttt{opt\_ndcg} and \(88\%\) for \texttt{opt\_pair}. Their first recommendations therefore continue to outperform the backend-specific MinFill plan in a similar proportion of test circuits. Identical aggregate percentages do not necessarily mean that the models win on the same circuits on both GPUs.

The fixed Ampere-trained rankings retain useful decision quality when assessed with Volta measurements, particularly when three candidate plans can be tested. The higher regret also shows that the backend change is not neutral and that the relative cost of a wrong recommendation can increase even when Top-3 and win rate remain high. This experiment provides evidence for zero-shot transfer between the two NVIDIA architectures considered here, not for general independence from the hardware and software backend.

\section{Conclusions}
\label{sec:conc}

The aim of this study was to select efficient contraction plans for GPU-based quantum circuit simulation without executing every candidate beforehand. We treated this as a ranking problem: for this application, identifying a good plan or a useful shortlist matters more than predicting exact runtimes.

The rankers use structural descriptors extracted directly from the sequence of contractions. These features capture properties related to computational load, parallelism, reductions, contraction geometry, and plan granularity, but require no preliminary execution or profiling on the GPU. The resulting model can therefore be placed between plan generation and execution as a low-cost selection stage.

The experiments show that this representation contains enough information to distinguish useful plans. On the locked in-distribution test, the NDCG-oriented model placed the fastest plan first in \(60\%\) of the circuits and within its first three recommendations in \(96\%\), with a mean \(\mathrm{Regret1}\) of \(0.0141\). It gave the best balance between Top-\(k\) accuracy and decision regret among the models considered. When the complete QFT-based family was excluded from model development, its Top-3 rate fell to \(62.9\%\). The drop is substantial, but the model still retained useful predictive information for a circuit family it had not seen during training or validation.

The cross-GPU results show a similar mixture of stability and change. Across the full dataset, Ampere and Volta selected the same fastest plan in \(84\%\) of the circuits, and their sets of three fastest plans had a mean overlap of \(89.6\%\). On the locked test set, the Ampere-trained NDCG model retained a Top-3 rate of \(92\%\) when evaluated with Volta measurements, although its regret increased. Structural properties of the plans therefore transfer to some extent across these two architectures, but the observed backend shift is too limited to support a general claim of hardware independence.

In practice, the method can reduce the candidate set to one plan or a short list that is then measured on the target GPU. Its usefulness still depends on the quality and diversity of the plans supplied to the ranker; the method selects among existing candidates rather than generating new contraction orders.

Our next step is to generate substantially larger and more varied candidate sets for each circuit, increasing both the dataset size and the range of plan structures represented in it. This will make it possible to study how ranking quality changes as the candidate pool grows. We also plan a stricter domain-shift experiment in which complete plan-generation methods or heuristic families are withheld from training and validation. In a later stage, we intend to include more distinct GPU backends and investigate rankers conditioned on descriptors of the target hardware and software stack, possibly supported by a small calibration set. That extension will depend on the portability of QXTools and its current CUDA-based environment.

\section*{Author Contributions}

Alfred M. Pastor: Software, Investigation, and Visualization. Maribel Castillo and Jose M. Badia: Supervision and Funding acquisition. All authors: Conceptualization, Methodology, Validation, Writing -- original draft, and Writing -- review \& editing.

All authors discussed the results, approved the final version of the manuscript, and take responsibility for its content.

ChatGPT (OpenAI) and Gemini (Google) were used as writing assistants for language editing, structural suggestions, and bibliographic and formal review. All scientific claims, numerical results, interpretations, software, experimental results, and final wording were verified and approved by the authors,
who take full responsibility for the manuscript.

\section*{Data and Code Availability}

The data and code supporting the results of this work are publicly available at
\url{https://github.com/alfred-miquel/LTR-Quantum-Contraction-Plans}.

The repository contains the circuit instances, candidate contraction plans, feature-extraction scripts, trained models, and analysis code used in this study.

\section*{Acknowledgments}

This work was supported by the research project PID2023-146569NB-C22 funded by MCIN/AEI/10.13039/501100011033 and ``ERDF A way of making Europe''.

\bibliographystyle{quantum}
\bibliography{biblio}

\onecolumn
\appendix

\section{Detailed cost model and feature-block definitions}
\label{app:feature_details}

This appendix provides the technical detail underlying the feature design summarized in Section~\ref{sec:feat_eng}. The aim is not to reproduce the internal cost model of a specific tensor-contraction library, but to define portable structural predictors that can be computed directly from a contraction plan before execution and that remain informative for GPU runtime. The starting point is the well-known observation that contraction cost is highly sensitive to path structure and that width-based reasoning, i.e., reasoning based on contraction width and related bottleneck-size proxies, while essential, is not sufficient to explain measured performance on current CPU/GPU tensor-network simulators~\cite{MaS08, Ogo19, DFG18, GrK21}.

\subsection{Why width alone is not sufficient}

Width-based theory explains why contraction order matters: simulation complexity is tightly linked to graph structure and, in particular, to treewidth-related notions of difficulty~\cite{MaS08, Ogo19, DFG18}. In the quantum-circuit setting, this connection is also borne out empirically: treewidth-oriented path search is often a strong practical proxy for contraction difficulty, although its effectiveness depends on the network family and on the optimization budget~\cite{GrK21, DFG18}. However, for feature engineering, width alone is not enough. Two plans with comparable leading width can still differ substantially in total work, local intermediate sizes, contraction-shape regularity, and parallel efficiency. This is especially important in current tensor-network simulators, where efficient execution relies on parallel tensor contractions on multicore CPUs and, increasingly, on GPUs~\cite{HZN21, VOA22, PGK24, PCB25, PBC25, BLA25}.

From the point of view of execution, the dominant factors are not only the computational burden of the bottleneck step, but also the amount of output-side parallelism, the reduction structure that governs reuse and arithmetic intensity, the geometry of the equivalent dense kernel, and the overhead induced by fragmented plans. This is consistent with Roofline-style reasoning, which makes explicit that achievable performance depends on the balance between arithmetic throughput and bandwidth-limited throughput rather than on FLOPs alone~\cite{WWP09}. Consequently, the cost model used to build the feature space must be actionable: it must be computable directly from the contraction plan before execution, yet expressive enough to reflect the structural properties that determine actual runtime on the target hardware.

\subsection{Local contraction model}

Consider again the \(i\)-th binary contraction step
\begin{equation*}
O_i = \mathrm{contract}_{S_i}(A_i,B_i),
\end{equation*}
where \(S_i\) is the set of shared indices eliminated in the contraction. For binary tensor networks,
\begin{align*}
k_i &= |S_i|,\\
m_i &= \mathrm{rank}(A_i)-k_i,\\
n_i &= \mathrm{rank}(B_i)-k_i,
\end{align*}
so that \(m_i\) and \(n_i\) count the free indices inherited from \(A_i\) and \(B_i\), respectively, and
\begin{equation*}
\mathrm{rank}(O_i)=m_i+n_i.
\end{equation*}

After a suitable permutation of modes, every pairwise contraction can be expressed in the GEMM-like form introduced in Eq.~\eqref{eq:gemm_view}. This abstraction is not merely pedagogical: it is exactly the type of decomposition exploited by modern tensor-contraction libraries, whether through explicit transpose--GEMM strategies, fused packing-and-contraction schemes, or native tensor kernels~\cite{SpB18, Mat18, Cor25, Qua25}. It therefore provides a natural basis for feature design.

\subsection{Step-level primitives and their interpretation}

At step level, we work with four primitives:
\begin{align*}
c_i &= m_i+n_i+k_i,\\
p_i &= m_i+n_i,\\
k_i &= |S_i|,\\
d_i &= \frac{|m_i-n_i|}{m_i+n_i+\varepsilon},
\qquad \varepsilon \ll 1.
\end{align*}

These four quantities form the irreducible basis of the feature space.

The first quantity, \(c_i\), is the work exponent. For binary networks, the equivalent FLOP count satisfies
\begin{equation}
\mathrm{FLOPs}_i=\Theta(2MNK)=\Theta(2^{1+c_i}),
\end{equation}
so \(c_i\) is the natural logarithmic proxy for local computational effort. It is the closest step-level quantity to the classical complexity surrogates traditionally used in contraction-path search. If one only wanted a combinatorial estimate of difficulty, \(c_i\) would be the natural first candidate.

The second quantity, \(p_i\), is the output-size exponent. Since \(|O_i|=2^{p_i}\), it has a dual interpretation. In the sequential regime it is a proxy for the immediate intermediate tensor size and hence for memory pressure, temporary workspace demand, and downstream data movement. In the parallel regime, because the output contains \(MN=2^{p_i}\) elements, it also acts as a proxy for exploitable data parallelism. This is why it is preferable to expose \(p_i\) explicitly rather than absorb it into a more opaque memory surrogate.

The third quantity, \(k_i\), is the reduction exponent. Although it appears inside \(c_i\), it captures a different structural aspect of the contraction. Two steps with the same \(c_i\) can have very different balances between output formation and inner reduction, and therefore very different reuse profiles. In Roofline terms, \(k_i\) influences whether the step is more likely to be compute-bound or memory-bound~\cite{WWP09}. This is also consistent with practice in GPU tensor contraction, where higher useful reuse and Tensor-Core-friendly structure are repeatedly identified as central to performance~\cite{PGK24, Cor25}.

The fourth quantity, \(d_i\), is a geometry descriptor. Values near zero correspond to roughly square contractions, whereas values near one indicate tall-and-skinny or short-and-wide cases. This distinction matters because optimized dense kernels are not shape-invariant: highly imbalanced contractions are more likely to suffer from poorer locality, weaker packing efficiency, less favorable stride structure, and lower sustained throughput~\cite{SpB18, Mat18, Cor25}. Thus, \(d_i\) acts as a compact surrogate for contraction geometry and mapping quality.

A minimal backend-agnostic proxy for data movement also follows from the GEMM-like form:
\begin{equation*}
\mathrm{Bytes}^{\min}_i = \Theta(MK+KN+MN).
\end{equation*}
This expression clarifies why separating \(p_i\) and \(k_i\) is meaningful: steps with similar FLOP counts can still differ substantially in output size, reduction depth, and therefore in the balance between computation and memory traffic.

\subsection{Plan-level aggregation}

A contraction plan \(\pi\) with \(S\) pairwise contractions is a variable-length sequence of local descriptors. To use it in a tabular learning model, we map it to a fixed-size vector through aggregations that preserve the physically relevant structure of the plan. Simple averages are rarely sufficient. In practical tensor-network contractions, total runtime is often dominated by one or a few expensive steps, while the long tail of small steps contributes mainly through overhead, intermediate materialization, or cache- and kernel-unfriendly fragmentation~\cite{GrK21, Ogo19}.

We therefore define the bottleneck exponent
\begin{align}
c_{\max} &= \max_i c_i, \notag \\
I_{\max} &= \{i:\, c_i=c_{\max}\},
\end{align}
and FLOP-proportional weights
\begin{equation*}
w_i = \frac{2^{c_i-c_{\max}}}{\sum_{\ell=1}^{S}2^{c_\ell-c_{\max}}}.
\end{equation*}
The subtraction of \(c_{\max}\) is only for numerical stability; it does not change the weighting.

The resulting features are grouped into blocks that capture distinct hypotheses about what limits runtime. These blocks are not arbitrary. They are designed to represent the dominant work region, the distribution of costs beyond the single bottleneck, and structural overhead effects that asymptotic complexity tends to ignore.

\subsection{Complexity and critical-load block}

The first aggregation block captures the overall computational burden of the plan together with the structure of its bottleneck region. It includes
\begin{align}
\mathrm{max\_cost}(\pi) &= c_{\max},\\
\mathrm{log2\_sum\_flops}(\pi) &= \log_2\!\left(\sum_{i=1}^{S}2^{c_i}\right),\\
\mathrm{topq\_mean\_cost}(\pi) &= \frac{1}{|I_q|}\sum_{i\in I_q} c_i,
\end{align}
where \(I_q\) denotes the set of indices corresponding to the \(\lceil qS\rceil\) largest values of \(c_i\), together with
\begin{align}
\mathrm{avg\_cost}(\pi) &= \frac{1}{S}\sum_{i=1}^{S} c_i,\\
\mathrm{std\_cost}(\pi) &= \mathrm{std}(c_i).
\end{align}

These descriptors are correlated but not redundant. \(\mathrm{max\_cost}\) is the direct descendant of width-based reasoning and tracks the dominant contraction. \(\mathrm{log2\_sum\_flops}\) moves beyond the leading term and measures total accumulated work. \(\mathrm{topq\_mean\_cost}\) captures whether the plan is dominated by one isolated critical step or by a broader near-critical tail. This distinction is important in practice because a plan with many almost-maximal steps can behave quite differently from a plan with one sharp peak and a cheap remainder. Finally, \(\mathrm{avg\_cost}\) and \(\mathrm{std\_cost}\) summarize the global level and heterogeneity of the cost profile: a high standard deviation indicates a very uneven plan, whereas a lower value indicates a more homogeneous contraction sequence.

\subsection{Sequence and granularity block}

The second block captures structural fragmentation:
\begin{align}
\mathrm{n\_steps}(\pi) &= S,\\
\mathrm{frac\_tiny\_steps}(\pi) &= \frac{1}{S}\sum_{i=1}^{S}\mathbf{1}\{c_i\le c_{\max}-\tau\},
\end{align}
where \(\tau\) is a fixed threshold, e.g.\ \(\tau=6\), so that a ``tiny'' step is at least \(2^6=64\) times smaller in FLOP scale than the bottleneck.

These variables matter because asymptotic work models ignore granularity. A plan can have a reasonable bottleneck and moderate total work, yet still perform poorly if it decomposes the contraction into too many very small operations. Such a plan tends to amplify loop and scheduling overhead, increase the number of intermediate tensors that must be created and consumed, and reduce the opportunity for sustained high-efficiency execution. In this sense, \(\mathrm{n\_steps}\) measures structural fragmentation, while \(\mathrm{frac\_tiny\_steps}\) identifies whether that fragmentation is concentrated in an extended tail of low-value operations.

\subsection{GPU-oriented refinement}

At this point, the sequential feature space already captures the main classical determinants of cost: bottleneck difficulty, total work, workload dispersion, and plan granularity. However, selecting a good plan for GPUs requires a more refined view of how each step maps to parallel hardware. Modern tensor-network simulators increasingly rely on multicore and GPU backends because pairwise contractions expose large amounts of data parallelism and can often be executed through GEMM-like kernels~\cite{HZN21, VOA22, PGK24, PCB25, PBC25}. On such hardware, runtime is not a monotone function of FLOPs. The Roofline model makes this explicit: achievable performance is bounded by the minimum of peak arithmetic throughput and bandwidth-limited throughput, so kernels with similar FLOP counts may fall into very different execution regimes depending on their arithmetic intensity~\cite{WWP09}.

For the local contraction in Eq.~\eqref{eq:gemm_view}, an intensity proxy is
\begin{equation}
I_i \approx \frac{2MNK}{MK+KN+MN}.
\end{equation}
For fixed \(M\) and \(N\), the dominant driver of \(I_i\) is \(K=2^{k_i}\). This is why \(k_i\) is promoted from an internal term in \(c_i\) to an explicit feature: it separates high-reuse, compute-friendly contractions from low-reuse, memory-sensitive ones. The variable \(p_i\) also acquires a second interpretation in the GPU regime. Since the output contains \(MN=2^{p_i}\) elements, \(p_i\) becomes a direct proxy for exploitable data parallelism: a large \(p_i\) offers many independent output elements to distribute across threads and thread blocks, whereas a very small \(p_i\) is more likely to underutilize the device even if the theoretical work exponent is large.

The geometry descriptor \(d_i\) is equally important. High-performance tensor libraries stress that memory layout and stride regularity are critical for performance. On CPUs, fused packing and transpose-avoidance are used to recover favorable access patterns~\cite{SpB18, Mat18}. On GPUs, the same issue appears as a requirement to align mode ordering across tensors, to keep batched modes in slow-varying dimensions, and to preserve a large stride-one extent whenever possible~\cite{Cor25}. This is why \(d_i\) matters: it is a compact surrogate for how square or skinny the effective \(M\times N\) geometry is, and therefore for how easily the step can be mapped to a high-throughput kernel.

\subsection{Parallelism-dynamics block}

This block summarizes how much useful output-side parallelism is present in the plan, globally and at the bottleneck:
\begin{align}
\mathrm{max\_out\_rank}(\pi) &= \max_i p_i,\\
\mathrm{avg\_out\_rank}(\pi) &= \frac{1}{S}\sum_i p_i,\\
\mathrm{costw\_out\_rank}(\pi) &= \sum_i w_i p_i,\\
\mathrm{p\_at\_max\_cost}(\pi) &= \frac{1}{|I_{\max}|}\sum_{i\in I_{\max}} p_i.
\end{align}

Each of these quantities answers a slightly different question. \(\mathrm{max\_out\_rank}\) tells us how much output parallelism is ever exposed by the plan and also acts as a rough proxy for the largest immediate intermediate tensor. \(\mathrm{avg\_out\_rank}\) summarizes the typical output scale of the plan, which is useful when discriminating between consistently parallel-friendly plans and those that only occasionally expose large outputs. \(\mathrm{costw\_out\_rank}\) is more selective: it asks whether the expensive part of the plan is also the parallel-rich part. This distinction is crucial because large output tensors in cheap steps do little to accelerate total runtime if the bottleneck itself remains poorly parallelized. Finally, \(\mathrm{p\_at\_max\_cost}\) isolates that exact issue by measuring output parallelism at the critical step.

\subsection{Reduction and arithmetic-intensity block}

This block isolates whether the heavy contractions of the plan are reduction-rich or reduction-poor:
\begin{align}
\mathrm{max\_red\_rank}(\pi) &= \max_i k_i,\\
\mathrm{costw\_red\_rank}(\pi) &= \sum_i w_i k_i,\\
\mathrm{k\_at\_max\_cost}(\pi) &= \frac{1}{|I_{\max}|}\sum_{i\in I_{\max}} k_i.
\end{align}

The purpose of this block is to make arithmetic intensity visible at plan level. \(\mathrm{max\_red\_rank}\) records the largest reduction dimension encountered anywhere in the path. On its own, that feature says little about total runtime, but it does indicate whether the plan contains potentially high-reuse contractions. \(\mathrm{costw\_red\_rank}\) is more informative because it measures the reduction-richness of the expensive region rather than of the path as a whole. \(\mathrm{k\_at\_max\_cost}\) goes one step further and characterizes the bottleneck directly. Plans with similar \(\mathrm{max\_cost}\) can differ markedly in \(\mathrm{k\_at\_max\_cost}\), and therefore in whether their dominant contraction is more likely to be compute-bound or memory-bound.

\subsection{Geometry and memory-access block}

This block summarizes contraction shape both globally and at the bottleneck:
\begin{align}
\mathrm{max\_asym}(\pi) &= \max_i d_i,\\
\mathrm{avg\_asym}(\pi) &= \frac{1}{S}\sum_i d_i,\\
\mathrm{costw\_asym}(\pi) &= \sum_i w_i d_i,\\
\mathrm{d\_at\_max\_cost}(\pi) &= \frac{1}{|I_{\max}|}\sum_{i\in I_{\max}} d_i.
\end{align}

These descriptors become particularly important when two plans have similar bottleneck FLOP exponents but very different effective GEMM geometries. \(\mathrm{max\_asym}\) identifies whether the plan ever enters an extremely unbalanced regime. \(\mathrm{avg\_asym}\) measures the overall geometric tendency of the path. \(\mathrm{costw\_asym}\) asks whether the expensive part of the plan is also the geometrically awkward part, and \(\mathrm{d\_at\_max\_cost}\) makes that question explicit at the bottleneck. Their common justification is that shape regularity is a first-order determinant of memory-access quality and kernel efficiency, particularly on GPUs where stride structure, packing efficiency, and Tensor Core utilization are strongly shape-dependent~\cite{SpB18, Mat18, Cor25}.

\subsection{Kernel-management and overhead block}

The previously introduced features \(\mathrm{n\_steps}\) and \(\mathrm{frac\_tiny\_steps}\) remain relevant here, but their interpretation sharpens on GPUs. Launch cost, plan creation, autotuning, and workspace management are non-negligible for fine-grained contractions, especially if the plan contains many steps far below the critical-load scale. Modern GPU libraries explicitly address these effects with planning stages, plan caches, JIT compilation, and resource reuse mechanisms~\cite{Cor25, Cor24}. Since such overheads are backend- and version-dependent, we do not expose them as explicit library-specific features; instead, we retain structural surrogates that remain portable across software stacks.

This design choice deserves emphasis. The goal is not to reproduce the private internal cost model of a specific library, but to define portable predictors of when library overhead is likely to matter. A high value of \(\mathrm{n\_steps}\) indicates a fragmented path that repeatedly re-enters the runtime system, while a high value of \(\mathrm{frac\_tiny\_steps}\) suggests a long tail of steps whose useful numerical work may be small compared with their execution-management cost. These are precisely the kinds of effects that asymptotic complexity neglects but measured runtime reveals.

\subsection{Final design principles and portability}

The final feature space is deliberately richer than a minimal cost summary. Several aggregated descriptors are correlated, but they encode distinct hypotheses about what actually limits runtime: a single catastrophic step, a broad near-critical tail, insufficient output parallelism in the dominant region, low reduction-driven reuse, unfavorable contraction geometry, or an accumulation of tiny-kernel overheads. Retaining this partial redundancy is preferable at the feature-design stage because it allows the downstream learning model, together with feature-importance analyses and ablations, to determine which structural mechanisms carry predictive signal on a given hardware/software stack.

Importantly, the feature vector is derived entirely from the contraction plan and therefore does not change when the same plan is executed on a different GPU. What changes across backends is the measured runtime, and therefore the supervision signal used for learning and evaluation. Finally, although the present work targets binary tensor networks induced by qubit circuits, the construction extends directly to heterogeneous bond dimensions by replacing rank counts with base-2 logarithms of the corresponding dimension products. For example, \(p_i\) becomes \(\log_2 |O_i|\), while \(M\), \(N\), and \(K\) in Eq.~\eqref{eq:gemm_view} become the true grouped dimensions of the equivalent matrix multiplication. The feature design is therefore specific to the circuit-simulation setting, but not tied to binary tensors in any essential way.

\end{document}